\documentclass[sigconf]{acmart}
\usepackage{multirow}
\usepackage{booktabs}
\usepackage{placeins}

\AtBeginDocument{%
  }

\setcopyright{none}                                
\renewcommand\footnotetextcopyrightpermission[1]{} 
\acmISBN{978-1-4503-XXXX-X/2018/06}

\begin{document}
\pagestyle{empty}

\title{Benchmarking Geospatial Foundation Models for Agriculture Applications}

\author{Zhuocheng Shang}
\affiliation{%
  \institution{University of California, Riverside}
  \city{Riverside}
  \country{USA}}
\email{zshan011@ucr.edu}

\author{Sanmay Das}
\affiliation{%
  \institution{University of California, Riverside}
  \city{Riverside}
  \country{USA}}
\email{sdas050@ucr.edu}

\author{Ahmed Eldawy}
\affiliation{%
  \institution{University of California, Riverside}
  \city{Riverside}
  \country{USA}}
\email{eldawy@ucr.edu}

\renewcommand{\shortauthors}{Shang et al.}

\begin{abstract}
Geospatial foundation models pretrained on satellite imagery promise broad generalization across remote sensing tasks and regions, but their geographic transferability has not been systematically tested, especially in agriculture applications. This paper presents a controlled benchmark that evaluates three models, Prithvi, SpectralGPT, and SatMAE, on multi-temporal crop segmentation and change detection across four U.S. states, Iowa, North Carolina, California, and Minnesota. By assigning each train, validation, and test split to a separate region, we measure how well each model transfers to land it has not seen. All three degrade sharply under regional distribution shift, predicting only the most common crops
while missing rare ones. We further find that fitting these models to a shared input format affects each one differently, which complicates direct architectural comparison. These results expose key limitations of
current geospatial foundation models for agriculture and point to region aware evaluation as a necessary standard.
\end{abstract}




\maketitle

\section{Introduction}
Agriculture occupies roughly 880 million acres in the United States, nearly 39\% of the national land area~\cite{usda2022census}. Managing this land at scale requires accurate crop type maps, which
identify what is grown and where. Such maps support food supply monitoring, yield estimation, water planning, and other agricultural decisions~\cite{joshi2025cropmapping}. However, crop mapping becomes difficult when models are applied across regions with different production systems, climates, and management constraints. For example,
annual row-crop production in the US Corn Belt differs sharply from
agriculture in California, where crop diversity, irrigation dependence,
and water stress create a substantially different mapping
setting~\cite{usda_ers_cropvalue, liu2022groundwater}. Such regional differences mean that the methods and models used in one agricultural setting may not be accurate in another, making geographic generalization a central challenge for national scale crop mapping.

Producing crop maps at national scale rules out manual ground surveys, which are labor intensive and cannot cover the area
involved~\cite{joshi2025cropmapping}. Satellite imagery removes this
bottleneck by capturing large regions in a single pass, and a growing body of work builds on it to monitor agriculture. Geospatial Foundation
Models (GeoFMs) such as Prithvi~\cite{jakubik2023foundation},
SatMAE~\cite{cong2022satmae}, and SpectralGPT~\cite{hong2024spectralgpt},
pretrained on large Earth observation datasets, promise generalizable representations for multi temporal crop segmentation and change detection with little task specific supervision. Whether they keep this
promise is unclear. The question that we would like to study is `\textit{Do these models generalize across geographies for agricultural crop mapping, or do reported results mainly reflect the quirks of a single dataset?}'

Existing benchmarks cannot settle this question, because each fails at least one of two requirements, geographic diversity beyond Europe and an evaluation that exposes generalization rather than concealing it.
PASTIS~\cite{garnot2021panoptic}, BreizhCrop~\cite{russwurm2019breizhcrops},
Sen4AgriNet~\cite{sykas2021sen4agrinet}, and
EuroCropsML~\cite{reuss2025eurocropsml} are confined to Europe. GEO-Bench~\cite{lacoste2023geobench} randomly subsamples chips, removing the spatial structure needed to test transfer to new regions. PANGAEA~\cite{marsocci2024pangaea} evaluates geographic transfer only on a single urban land cover dataset, not crop mapping, and PhilEO-Bench~\cite{fibaek2024phileo} does not target agriculture. As a result, whether GeoFMs transfer across U.S. geographies for crop mapping
remains untested.

We close this gap with a benchmark that evaluates
Prithvi~\cite{jakubik2023foundation},
SpectralGPT~\cite{hong2024spectralgpt}, and
SatMAE~\cite{cong2022satmae} on multi-temporal crop segmentation and change detection using Sentinel-2 imagery~\cite{sentinel2_eoportal}
across four climatically diverse U.S. states, Iowa, North Carolina, California, and Minnesota. For each model, we use separate non-overlapping regions for training, validation, and testing, so that
performance reflects transfer to unseen geographic areas rather than spatial overlap between splits. This design lets us measure how well each model generalizes across agricultural regions. Our results show
that \textbf{all three models degrade sharply when tested on a new region}. Under this regional distribution shift, the models often concentrate on dominant crop classes while failing to detect minority
crops. We further find that adapting these models to a shared input configuration affects their pretrained representations unequally, which complicates direct architectural comparison. These findings suggest
that region-aware evaluation and input-preserving adaptation are necessary steps toward geospatial foundation models that transfer reliably across agricultural regions.

\section{Related Work}
\textbf{Geospatial foundation models} build large neural networks trained on Earth observation imagery to learn transferable representations for remote sensing tasks.
Prithvi~\cite{jakubik2023foundation} uses multitemporal HLS imagery and has been evaluated on Cropland Data Layer (CDL) for crop segmentation. SatMAE~\cite{cong2022satmae}
learns spectral group embeddings from functional map of the world (fMoW) Sentinel imagery. SpectralGPT~\cite{hong2024spectralgpt} emphasizes spectral structure for understanding remote sensing. These models introduce different temporal
and spectral inductive biases, but existing evaluations do not isolate geographic transfer as the main variable. In particular, they either use pooled training and test samples or target broader land cover and land use tasks rather than controlled cross region crop mapping. We evaluate these models under controlled geographic splits across diverse U.S. agricultural regions.

\textbf{Crop mapping benchmarks.}
Several datasets support crop type mapping from satellite time series.
PASTIS~\cite{garnot2021panoptic},
BreizhCrop~\cite{russwurm2019breizhcrops},
Sen4AgriNet~\cite{sykas2021sen4agrinet}, and
EuroCropsML~\cite{reuss2025eurocropsml} provide labeled imagery for segmentation and classification, but all four are confined to Europe and
therefore cannot reveal whether a model trained in one climate transfers to another. Their labels are reported by farmers, which introduces
systematic error that propagates into both training and evaluation. As a result, these benchmarks measure accuracy within a region rather than
transfer across regions.

\textbf{Benchmarking and evaluation of foundation models.}
A separate line of work evaluates how well geospatial foundation models generalize. GEO-Bench~\cite{lacoste2023geobench} assembles many tasks
but randomly subsamples image chips, removing the spatial structure needed to test transfer to new regions. PANGAEA~\cite{marsocci2024pangaea} reports
that supervised baselines often match or exceed GeoFMs, and includes a regional transfer test, but on a single urban land cover dataset rather than crop mapping. PhilEO-Bench~\cite{fibaek2024phileo} evaluates only
on Sentinel-2 data and does not target agriculture. None evaluates U.S. crop mapping or tests transfer across controlled, climatically diverse
regions, so whether GeoFMs transfer across U.S. agricultural geographies remains open. Our benchmark fills this gap through strict geographic splits that isolate generalization as the variable under test.
\section{Datasets and Experimental Setup}
Throughout this paper, train, validation, and test refer to the downstream supervised fine tuning and evaluation splits, not to the original foundation model pretraining stage. Each GeoFM starts from its publicly released pretrained weights, is fine tuned on labeled chips from
the training region, selected on the validation region, and evaluated only on the held out test region.
\textbf{Study regions.}
For each studied state, we define three non overlapping contiguous regions of
800--1{,}300~km$^{2}$ and assign one each to train, validation, and test as in Table~\ref{tab:study_regions}. This allows us to test each model on an area that it did not see during training to evaluate their geographical transferability. The four states span contrasting agricultural conditions. Iowa and Minnesota cover Corn Belt row crops,
North Carolina covers coastal and muck soil agriculture, and California covers perennial, irrigation dependent agriculture, the most severe distribution shift.

\begin{table}[t]
\centering
\caption{Study regions for this benchmark.}
\label{tab:study_regions}
\begin{tabular}{llll}
\toprule
\textbf{State} & \textbf{Train} & \textbf{Val.} & \textbf{Test} \\
\midrule
Iowa           & Ames        & Cedar Rapids & Sheldon--Sibley \\
North Carolina & Rocky Mount & Clinton      & Belhaven \\
California     & Rio Vista   & Merced       & Bakersfield \\
Minnesota      & Crookston     & Willmar    & Rochester \\
\bottomrule
\end{tabular}
\end{table}

\begin{figure}[t]
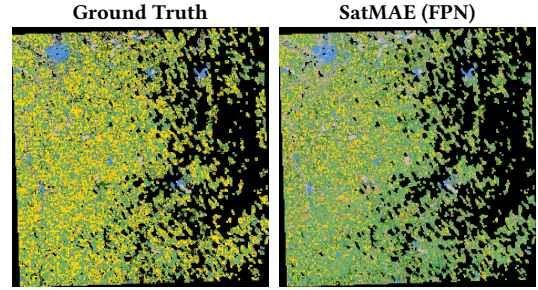

\centering
\setlength{\tabcolsep}{2pt}
\begin{tabular}{cc}
\small \textbf{Ground Truth} & \small \textbf{SatMAE (FPN)} \\
\includegraphics[width=0.4\columnwidth]{SouthMN_GT_colored.png} &
\includegraphics[width=0.4\columnwidth]{SouthMN_SatMAE_FPN_colored.png} \\
\end{tabular}
\caption{SatMAE (FPN) segmentation on the Minnesota test region
(mIoU 8.11). Corn is yellow and Soybean dark green.}
\label{fig:satmae_pred}
\end{figure}
\begin{figure}[t]
\centering
\includegraphics[width=0.75\columnwidth]{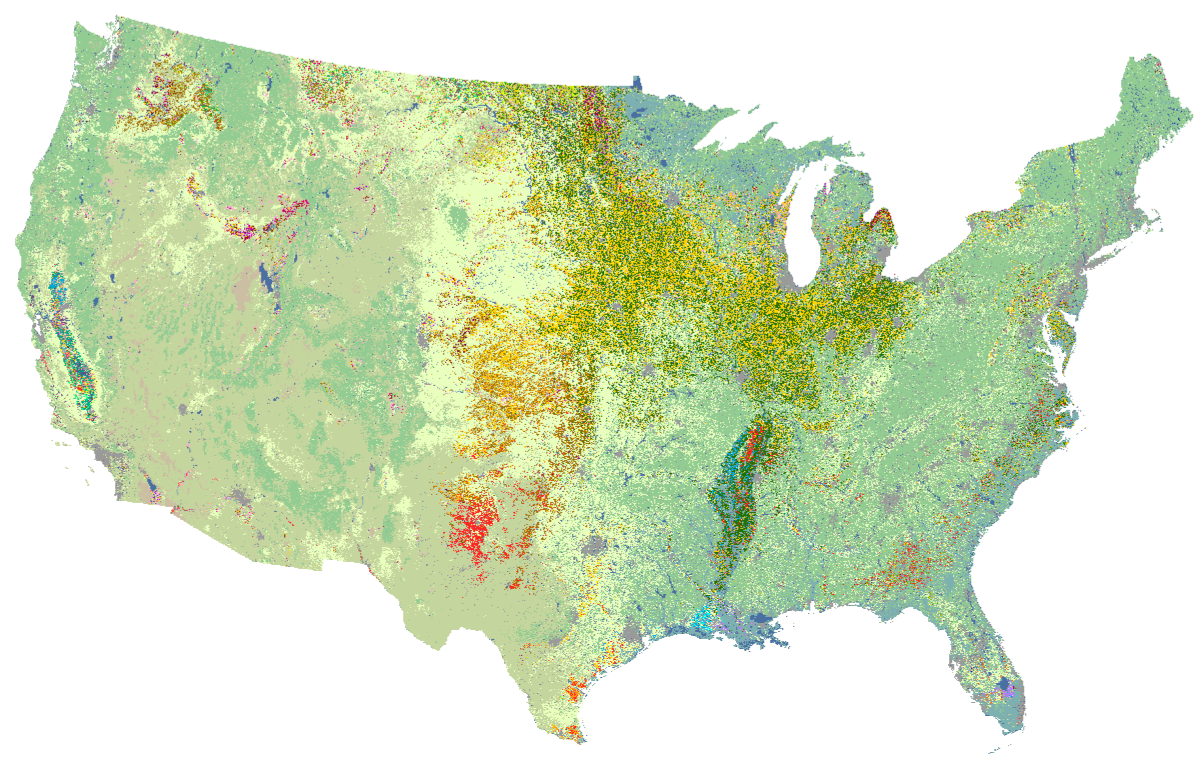}
\caption{USDA Cropland Data Layer ground truth, with the crop and land cover classes used as labels.}
\label{fig:cdl_conus}
\end{figure}

\textbf{Imagery and labels.}
We use Sentinel-2 Level-2A imagery from the Copernicus Data Space Ecosystem~\cite{cdse_annual_report} for the 2024 growing season. For each location, we select six bands, B2, B3, B4, B8A, B11, and B12~\cite{sentinel2_bands}, at early, mid, and late season time points. Scenes are cloud-filtered at acquisition and reprojected to EPSG:5070 at 10~m resolution, with all bands resampled to the common grid. We stack the six bands across the three time points to form an 18-band input. We then create image chips of size $224 \times 224$ for Prithvi, $128 \times 128$ for SpectralGPT, and $96 \times 96$ for SatMAE to match each model's expected input size. Labels are derived from the 2024 USDA Cropland Data Layer~\cite{usda_cdl}, mapped to 13 classes, and used for both tasks. For change detection, a pixel is labeled as changed if its CDL class differs between 2023 and 2024; pixels with missing CDL data in either year are excluded.

\textbf{Models.}
We evaluate three GeoFMs with native multispectral support beyond RGB, which excludes Scale-MAE~\cite{reed2023scale} and
SAM~\cite{osco2023segment, gurav2023sam}. We retain
Prithvi~\cite{jakubik2023foundation}, SpectralGPT~\cite{hong2024spectralgpt},
and SatMAE~\cite{cong2022satmae}, which differ in whether they prioritize
temporal or spectral structure, each with its own pretrained encoder as in Table~\ref{tab:model_config}. Since PSANet gave near zero mIoU with the
SatMAE backbone~\cite{cong2022satmae}, we use FPN heads for SatMAE and SpectralGPT and an FCN head for Prithvi. Because the head varies with the
encoder, we treat absolute cross model differences with caution and focus on each model across regions. For change detection, all three use a Siamese encoder with shared weights and an FPN style head.

\textbf{Inference.}
Segmentation uses TerraTorch tiled inference~\cite{gomes2025terratorch}
with reflect padding, cosine tapered blending, and an 8 pixel border discard. We report mIoU and per class IoU for segmentation, and F1 and
IoU\textsubscript{change} for change detection following the SpectralGPT protocol~\cite{hong2024spectralgpt}. All runs use a single NVIDIA A100.

\begin{table}[t]
\centering
\caption{Segmentation configuration. Loss is cross entropy. SatMAE uses SGD with momentum 0.9, encoder LR $1\times10^{-3}$,
weight decay $1\times10^{-4}$, and effective batch 128.}
\label{tab:model_config}
\small
\begin{tabular}{@{}llll@{}}
\toprule
& \textbf{Prithvi} & \textbf{SpectralGPT} & \textbf{SatMAE} \\
\midrule
Backbone    & ViT MAE            & 3D spectral MAE  & GroupChannels ViT-L \\
Pretraining & HLS                & fMoW-S2          & fMoW-S2 \\
Seg.\ head  & ConvT + FCN        & FPN              & FPN \\
Optimizer   & Adam               & AdamW            & SGD \\
LR          & $1.5\times10^{-5}$ & $1\times10^{-4}$ & $1\times10^{-2}$ \\
Epochs      & 80                 & 250              & 100 \\
Loss        & weighted CE        & CE               & CE \\
\bottomrule
\end{tabular}
\end{table}
\section{Results}
\subsection{Segmentation}
\begin{table}[t]
\centering
\caption{Change detection F1 (\%) across four U.S.\ states.}
\label{tab:cd_results}
\setlength{\tabcolsep}{4pt}
\begin{tabular}{@{}lcccc@{}}
\toprule
\textbf{Model} & \textbf{IA} & \textbf{NC} & \textbf{CA} & \textbf{MN} \\
\midrule
SpectralGPT & \textbf{68.20} & 48.81 & 18.73 & \textbf{86.05} \\
Prithvi     & 55.03 & 53.80 & 23.64 & 83.26 \\
SatMAE      & 41.10 & \textbf{55.65} & \textbf{27.00} & 67.89 \\
\bottomrule
\end{tabular}
\end{table}
\begin{table}[t]
\centering
\caption{Per-class segmentation IoU (\%) for all three models on Iowa
and Minnesota. NaN = class absent.}
\label{tab:iou_satmae_two}
\scriptsize
\setlength{\tabcolsep}{2.5pt}
\resizebox{\columnwidth}{!}{%
\begin{tabular}{@{}lcccccc@{}}
\toprule
\multirow{2}{*}{\textbf{Class}} &
\multicolumn{3}{c}{\textbf{Iowa}} &
\multicolumn{3}{c}{\textbf{Minnesota}} \\
\cmidrule(lr){2-4}\cmidrule(lr){5-7}
& \textbf{Pri} & \textbf{SGPT} & \textbf{SatMAE}
& \textbf{Pri} & \textbf{SGPT} & \textbf{SatMAE} \\
\midrule
\multicolumn{7}{l}{\textit{Crops}} \\
Corn         & 33.69 & 7.99 & \textbf{92.10} & 4.04 & \textbf{46.86} & 14.15 \\
Soybean      & 38.41 & 26.70 & \textbf{92.41} & 17.27 & \textbf{62.14} & 3.15 \\
Winter Wheat & 0.00 & 0.00 & 0.00 & 0.00 & 0.00 & 0.00 \\
Alfalfa      & 0.00 & 30.62 & \textbf{63.38} & 0.12 & \textbf{9.31} & 0.73 \\
Fallow/Idle  & 0.00 & 0.00 & \textbf{26.21} & 0.00 & \textbf{23.57} & NaN \\
Cotton       & NaN & NaN & NaN & NaN & NaN & 0.00 \\
Sorghum      & 0.00 & 0.00 & 0.00 & 0.00 & 0.00 & 0.00 \\
\midrule
\multicolumn{7}{l}{\textit{Land cover}} \\
Natural Veg.     & 0.00 & 41.86 & \textbf{78.19} & 0.01 & \textbf{14.94} & 8.61 \\
Forest           & 0.73 & 8.62 & \textbf{79.62} & 2.16 & \textbf{44.38} & 32.30 \\
Wetlands         & 0.00 & 2.03 & \textbf{44.93} & 1.69 & \textbf{14.08} & 9.27 \\
Developed/Barren & 1.51 & 53.35 & \textbf{66.01} & 0.32 & \textbf{52.83} & 19.83 \\
Open Water       & 6.86 & 39.44 & \textbf{67.71} & 0.76 & \textbf{75.35} & 0.00 \\
Other            & 0.00 & 0.08 & \textbf{29.72} & 0.93 & \textbf{8.35} & NaN \\
\midrule
\textbf{mIoU (all)}   & 6.77 & 17.59 & \textbf{53.36} & 2.28 & \textbf{29.32} & 8.11 \\
\textbf{mIoU (crops)} & \textbf{12.02} & 10.94 & 45.68 & 3.57 & \textbf{23.65} & 3.20 \\
\bottomrule
\end{tabular}}
\end{table}
\begin{table*}[!t]
\centering
\caption{Per-class IoU (\%) for Prithvi and SpectralGPT, grouped into
crop and land cover classes. Bold marks the higher of the two models in
each state, omitted where both fall below 1.0. mIoU (all) averages all
13 classes; mIoU (crops) averages the seven crop classes only. NaN =
class absent in that region.}
\label{tab:iou_four_states}
\begin{tabular}{lcccccccccc}
\toprule
\multirow{2}{*}{\textbf{Class}} &
\multicolumn{2}{c}{\textbf{Iowa}} &
\multicolumn{2}{c}{\textbf{North Carolina}} &
\multicolumn{2}{c}{\textbf{California}} &
\multicolumn{2}{c}{\textbf{Minnesota}} &
\multicolumn{2}{c}{\textbf{mIoU}} \\
\cmidrule(lr){2-3}\cmidrule(lr){4-5}\cmidrule(lr){6-7}\cmidrule(lr){8-9}\cmidrule(lr){10-11}
& \textbf{Pri} & \textbf{SGPT}
& \textbf{Pri} & \textbf{SGPT}
& \textbf{Pri} & \textbf{SGPT}
& \textbf{Pri} & \textbf{SGPT}
& \textbf{Pri} & \textbf{SGPT} \\
\midrule
\multicolumn{11}{l}{\textit{Crops}} \\
\midrule
Corn         & \textbf{33.69} & 7.99  & 9.58 & \textbf{48.83} & 0.00 & 0.75 & 4.04 & \textbf{46.86} & 11.83 & \textbf{26.11} \\
Soybean      & \textbf{38.41} & 26.70 & 6.12 & \textbf{59.84} & NaN  & NaN  & 17.27 & \textbf{62.14} & 20.60 & \textbf{49.56} \\
Winter Wheat & 0.00 & 0.00 & 0.00 & 0.00 & 0.00 & 0.00 & 0.00 & 0.00 & 0.00 & 0.00 \\
Alfalfa      & 0.00 & \textbf{30.62} & 0.00 & 0.00 & 0.09 & \textbf{41.08} & 0.12 & \textbf{9.31} & 0.05 & \textbf{20.25} \\
Fallow/Idle  & 0.00 & 0.00 & 0.00 & 0.00 & 0.00 & \textbf{2.80} & 0.00 & \textbf{23.57} & 0.00 & \textbf{6.74} \\
Cotton       & NaN  & NaN  & 2.35 & \textbf{66.90} & 0.00 & 0.00 & NaN & NaN & 1.18 & \textbf{33.45} \\
Sorghum      & 0.00 & 0.00 & 0.00 & \textbf{4.94} & 0.00 & 0.00 & 0.00 & 0.00 & 0.00 & \textbf{1.24} \\
\midrule
\multicolumn{11}{l}{\textit{Land cover}} \\
\midrule
Natural Veg.     & 0.00 & \textbf{41.86} & 11.10 & \textbf{32.06} & 1.46 & \textbf{74.86} & 0.01 & \textbf{14.94} & 3.14 & \textbf{40.93} \\
Forest           & 0.73 & \textbf{8.62} & 65.42 & \textbf{88.43} & \textbf{12.98} & 2.74 & 2.16 & \textbf{44.38} & 20.32 & \textbf{36.04} \\
Wetlands         & 0.00 & \textbf{2.03} & 0.00 & 0.00 & 0.00 & 0.00 & 1.69 & \textbf{14.08} & 0.42 & \textbf{4.15} \\
Developed/Barren & 1.51 & \textbf{53.35} & 3.22 & \textbf{56.43} & 6.95 & \textbf{23.91} & 0.32 & \textbf{52.83} & 3.00 & \textbf{46.63} \\
Open Water       & 6.86 & \textbf{39.44} & 0.13 & \textbf{90.26} & 11.17 & \textbf{34.76} & 0.76 & \textbf{75.35} & 4.73 & \textbf{59.95} \\
Other            & 0.00 & 0.00 & 4.51 & \textbf{64.75} & 20.35 & \textbf{62.96} & 0.93 & \textbf{8.35} & 6.45 & \textbf{34.04} \\
\midrule
\textbf{mIoU (all)}   & 6.77 & \textbf{17.59} & 8.65 & \textbf{44.20} & 4.83 & \textbf{24.03} & 2.28 & \textbf{29.32} & 5.63 & \textbf{28.79} \\
\textbf{mIoU (crops)} & \textbf{12.02} & 10.94 & 2.58 & \textbf{25.95} & 0.02 & \textbf{7.45} & 3.57 & \textbf{23.65} & 4.55 & \textbf{16.97} \\
\bottomrule
\end{tabular}
\end{table*}
Each model takes an 18 band multitemporal chip and produces a dense per pixel prediction over the 13 CDL classes, trained on the training region
of each state, selected on the validation region, and evaluated on the held out test region. Prithvi and SpectralGPT are evaluated on all four
states as in Table~\ref{tab:iou_four_states}. SatMAE is evaluated on Iowa and
Minnesota, with North Carolina and California left to future work as in Table~\ref{tab:iou_satmae_two}.

We focus on crop classes, since the aggregate mIoU is inflated by stable land cover classes such as Forest, Open Water, and Developed/Barren that are easy to map and change little across regions. On crops, SpectralGPT
is the strongest of the fully evaluated models but still degrades sharply with regional shift, from 25.95 crop mIoU in North Carolina to 7.45 in
California. Prithvi reaches usable crop IoU only on Corn and Soybean in Iowa, and collapses to near zero elsewhere, including California. Minority crops such as Winter Wheat and Sorghum are
near zero for nearly all models and regions, effectively unrecognized under distribution shift. SatMAE shows a large apparent region dependence, from 45.68 crop mIoU in Iowa to 3.20 in Minnesota. Because this gap is unusually large, we are
auditing the SatMAE evaluation for consistent class index handling before interpreting it as a model level effect.

\subsection{Change Detection}
Each model takes a bitemporal 2023--2024 image pair and predicts a binary mask of pixels whose CDL class changed between the two years, using a Siamese encoder with shared weights. We measure F1 on the
changed class over the held out test region as in Table~\ref{tab:cd_results}.

Change detection shows the same regional pattern as segmentation. All three models peak in Minnesota, with F1 above 83 for SpectralGPT and
Prithvi, and bottom out in California, where F1 falls to 18.73 for SpectralGPT and 27.00 for SatMAE. The difficulty ordering of regions is consistent across models, so it reflects the region rather than the
architecture. No model leads everywhere: SpectralGPT is strongest in Iowa and Minnesota, while SatMAE and Prithvi pull ahead in North Carolina and California, the two hardest regions.

\section{Discussion and Conclusion}
The finding is that geography matters as much as architecture. For the two fully evaluated models, Prithvi and SpectralGPT, accuracy changes substantially across held out test areas, even though the splits are geographically disjoint within each state.
The effect falls hardest on crops. Land cover classes such as forest and water are easier, especially for SpectralGPT, so a pooled mIoU is carried by them and overstates how well the crops themselves are mapped.

Comparing the models fairly is difficult because they do not begin from the same input assumptions. Our chips carry 18 channels, six bands over three timesteps. Prithvi was pretrained on this exact structure and keeps
its input layer intact, while SpectralGPT and SatMAE were not. SpectralGPT
discards its pretrained patch embedding when the shape changes, and SatMAE averages and expands its existing weights. Yet even Prithvi, with the
closest match, does not generalize well, so a compatible input is not by itself enough.

The broader lesson is that strong numbers on one dataset do not promise the same elsewhere. Crop mapping benchmarks should test across separated
geographies and report crop specific scores, not a single aggregate.
Three limitations remain. Each model was run once, so we report observed differences rather than significance. SatMAE segmentation covers only Iowa and Minnesota, and the large gap between them needs further checks
before we draw conclusions. And the models differ in decoder and input adaptation, so finishing the SatMAE runs, equalizing the input setup, and adding repeated seeds are the clear next steps.



\FloatBarrier
\bibliographystyle{ACM-Reference-Format}
\bibliography{references}

\end{document}